\documentclass[runningheads]{llncs}
\usepackage[T1]{fontenc}
\usepackage{amsmath, amssymb}
\usepackage{graphicx,verbatim}
\usepackage{booktabs}
\usepackage{multirow}
\usepackage{caption}
\usepackage{cite}
\usepackage{bbding}
\begin{document}
\title{Physics-Driven Zero-Shot MRI Reconstruction with Non-local Image Priors}
%
\author{Lingtong Zhang, Wenlei Li, Mu He, Li Xiao, Yang Ji\inst{(}\Envelope\inst{)}}  
\authorrunning{L. Zhang et al.}
\institute{School of Information Science and Technology,\\ University of Science and Technology of China, Hefei, China \\
    \email{\{zhanglingtong, wenlei.li, muhe\}@mail.ustc.edu.cn, xiaoli11@ustc.edu.cn}\\
    \email{jiyang@ustc.edu.cn}}
\maketitle              
\begin{abstract}
Zero-Shot Self-Supervised Learning (ZS-SSL) has emerged as a promising paradigm for accelerated Magnetic Resonance Imaging (MRI) reconstruction, eliminating the reliance on fully-sampled external datasets. However, learning solely from a single under-sampled scan suffers from supervision scarcity and optimization instability, often leading to overfitting or artifacts. To address these challenges, we propose a robust physics-driven ZS-SSL framework that synergizes physical consistency with image-domain non-local priors. Our method introduces three core innovations: (1) a Coil Sensitivity Map (CSM)-Guided Dynamic Repository, which stabilizes the training trajectory by filtering physically inconsistent artifacts based on coil sensitivity constraints; (2) a SPIRiT-based regularization, which enforces k-space self-consistency via a learned correlation kernel and stochastic masking; (3) a Non-Local Self-Similarity
(NSS) Pixel Bank, which leverages the high-fidelity reference established by the former modules to explicitly mine non-local anatomical similarities, thereby augmenting supervision in the image domain. Extensive experiments on the FastMRI dataset demonstrate that our approach achieves state-of-the-art performance, particularly under high acceleration factors, effectively bridging the gap between zero-shot learning and supervised methods. The code is available at \url{https://github.com/Zolento/NS-SSL}. 

\end{abstract}
\section{Introduction}
Magnetic Resonance Imaging (MRI) is a fundamental diagnostic tool offering superior soft-tissue contrast without ionizing radiation. However, slow acquisition limits patient throughput and induces motion artifacts. Consequently, accelerating MRI via under-sampled reconstruction has become a key research focus. While Deep Learning (DL) surpasses traditional Compressed Sensing (CS) and Parallel Imaging (PI), most state-of-the-art methods rely on supervised training with fully-sampled datasets. Acquiring such ground truth is often clinically infeasible, and supervised models frequently suffer from poor generalization across distributions.

To reduce reliance on ground truth, Self-Supervised Learning (SSL) has emerged. Yet, dataset-based SSL methods \cite{ssdu,spicer} still require a large collection of under-sampled data for training, risking distribution shifts due to varying anatomies, scanners, or sequences. To achieve true independence from external data, Zero-Shot SSL (ZS-SSL) \cite{zs-ssl} was proposed to learn reconstruction mappings solely from a single scan.

Despite its success in removing artifacts without external supervision, existing ZS-SSL frameworks face distinct limitations. First, conventional ZS-SSL primarily exploits self-supervision within the k-space domain, largely overlooking the rich, non-local self-similarity priors available in the image domain. Second, training on a single subject carries an inherent risk of overfitting to noise or specific artifacts. While validation-based early stopping helps, the lack of robust, physics-driven regularization can lead to suboptimal convergence.  

Robust MRI reconstruction fundamentally relies on physics constraints. Traditional methods like SENSE \cite{pruessmann1999sense} and SPIRiT \cite{lustig2010spirit} exploit coil sensitivities and k-space self-consistency to provide stable, shift-invariant inductive biases, overcoming the generalization limits of pure data-driven DL. Furthermore, the inherent non-local self-similarity (NSS) ubiquitous in images (e.g., repetitive structures and textures) \cite{huang2021neighbor2neighbor,ma2025pixel2pixel,mansour2023zero} can serve as a powerful regularizer for ZS-SSL. By acting as pseudo-samples, NSS could effectively augment the scarce supervision signals inherent in single-scan zero-shot learning.

Based on the above analysis, we propose a robust ZS-SSL framework synergizing physics-driven consistency with image-domain self-similarity. Firstly, we introduce the Coil Sensitivity Map (CSM)-Guided Dynamic Repository augmented by SPIRiT-based regularization. This module dynamically filters artifacts and enforces multi-coil physical consistency, establishing a high-fidelity pseudo-label during training. Secondly, leveraging this stable reference, we introduce an NSS Pixel Bank to increase training samples. Unlike prior works that strictly partition k-space, this module explicitly mines non-local patch similarities within the image domain, effectively synthesizing augmented training samples from the single subject itself. Our main contributions are summarized as follows:
\begin{enumerate}
\item We design a CSM-Guided Dynamic Repository to obtain high-confidence k-space points based on coil sensitivity consistency, ensuring robust data fidelity and providing stable targets for ensemble fusion.
\item We introduce a SPIRiT-based regularization operation to enforce k-space self-consistency. By integrating a learned kernel with a stochastic masking strategy, this term imposes strict physical constraints while preventing the network from overfitting to kernel-specific artifacts.
\item We propose an NSS Pixel Bank to alleviate the supervision scarcity inherent in zero-shot learning. By explicitly mining repetitive similar structures, this mechanism constructs high-fidelity image-domain priors to augment training without external data.
\end{enumerate}
\begin{figure}[ht]
    \centering
    \includegraphics[width=1\textwidth]{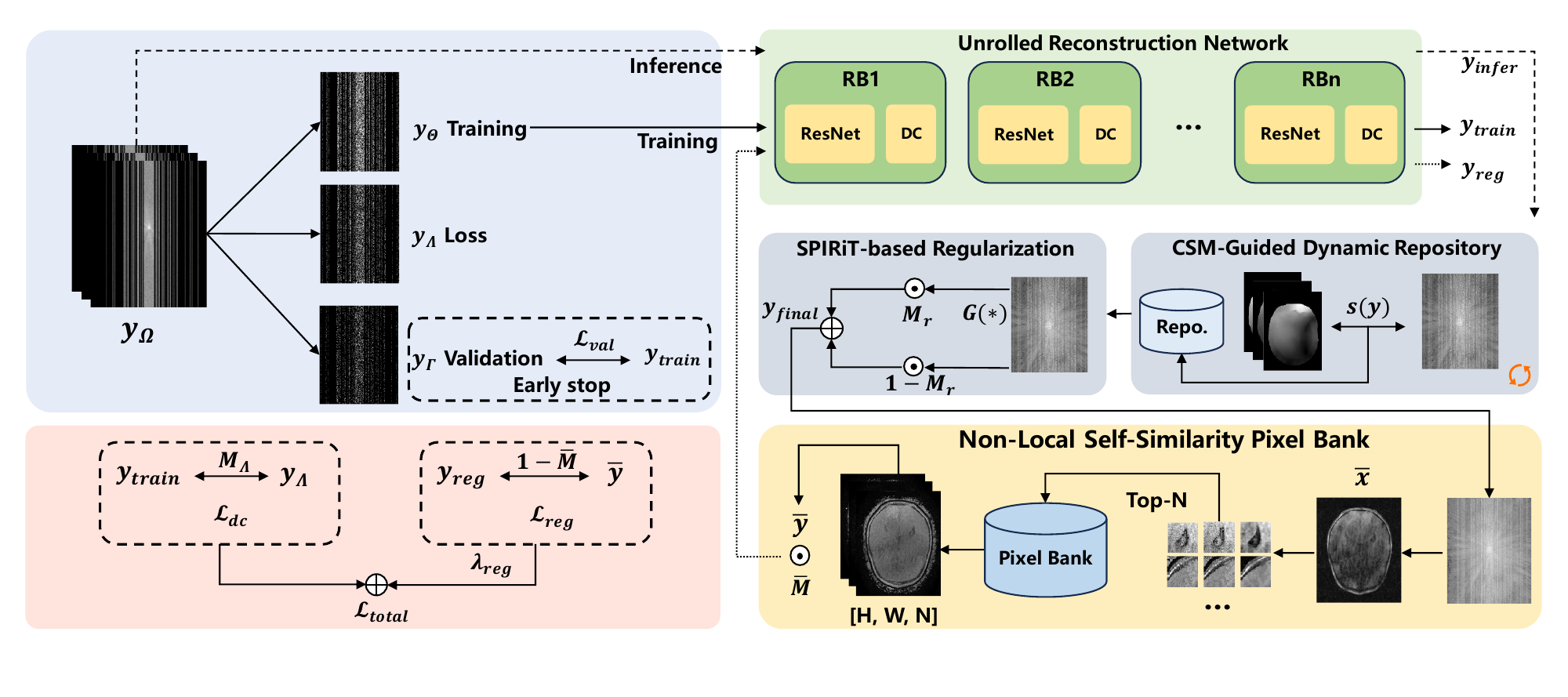}    
    \caption{Overview of our proposed method. For the sake of simplicity, the disjoint training pairs $\{(\Theta_k, \Lambda_k)\}_{k=1}^K$ are omitted. In the upper right corner, different line segments represent one-to-one corresponding input and output combinations.}
    \label{fig:overview}
\end{figure}
\section{Method}
\label{sec:method}
As shown in Fig. \ref{fig:overview}, our framework addresses supervision scarcity by integrating the CSM-Guided Dynamic Repository and the SPIRiT-based regularization to enforce physical consistency. These modules generate stable pseudo-labels to construct an NSS Pixel Bank, which augments training by mining image-domain anatomical priors for robust reconstruction.
\subsection{Self-Supervised MRI Reconstruction}
\label{subsec:ssl}
Let $y \in \mathbb{C}^{C \times H \times W}$ be the k-space measurement of $C$ coils, and $x\in \mathbb{C}^{H \times W}$ denote the image to recover. The forward model is:
\begin{equation}
y=M\mathcal{FS}x+n,\quad n \sim \mathcal{N}(0, \sigma^2_y I),
\end{equation}
where $M$ is the under-sampling mask, $\mathcal{F}$ is the Fourier Transform, and $\mathcal{S}$ denotes the CSMs. Following the standard ZS-SSL framework, we partition the original mask $\Omega$ into a validation mask $\Gamma$ for early stopping, and $\Omega' = \Omega \setminus \Gamma$ for training. To enable self-supervision without external data, $\Omega'$ is partitioned $K$ times into disjoint training ($\Theta_k$) and loss ($\Lambda_k$) masks, such that $\Omega' = \Theta_k \sqcup \Lambda_k$ with a uniform selection ratio $|\Lambda_k|/|\Omega'| = \rho$. The network $f$ with parameters $\theta$ is updated by minimizing the data consistency loss averaged over these $K$ partitions:
\begin{equation}
   \mathcal{L}_{\text{dc}} = \frac{1}{K} \sum_{k=1}^{K} \mathcal{L}\big(y_{\Lambda_k}, M_{\Lambda_k}(f(y_{\Theta_k}, M_{\Theta_k}; \theta))\big).
\end{equation}
Concurrently, the generalization capability is monitored via the validation loss: $\mathcal{L}_{\text{val}} = \mathcal{L}\big(y_\Gamma, M_\Gamma(f(y_{\Omega'}, M_{\Omega'}; \theta))\big)$.

\subsection{CSM-Guided Dynamic Repository}
\label{sec:repo}
To mitigate stochastic fluctuations inherent in single-instance training and filter out physically inconsistent artifacts, we employ the CSM-Guided Dynamic Repository, which continuously archives high-fidelity k-space components based on their agreement with CSMs. For each $k$ in epoch $t$, given the estimated full k-space $y_{\text{infer}}^{(t)}$ of $y_\Omega$ from the network, the repository $R$ is updated element-wise:
\begin{equation}
R_{t+1}[h,w] = 
\begin{cases}
y_{\text{infer}}^{(t)}[h,w], & \text{if } s(y_{\text{infer}}^{(t)}[h,w]) < s(R_t[h,w]), \\
R_t[h,w], & \text{otherwise},
\end{cases}
\end{equation}
where the CSM-guided consistency score $s(y)$ measures the deviation from the SENSE model:
\begin{equation}
s(y) = \|y - \mathcal{F}\mathcal{S} \mathcal{S}^{H} \mathcal{F}^{H}(y)\|_2.
\end{equation}

\subsection{SPIRiT-based Regularization}
\label{sec:spiritsfuse}
SPIRiT \cite{lustig2010spirit} leverages the self-consistency property of multi-coil MRI, positing that each point in the k-space can be synthesized as a linear combination of its neighboring points across all coils due to inter-coil correlations. We calibrate a SPIRiT kernel $\mathbf{G} \in \mathbb{C}^{C \times C \times \kappa \times \kappa}$ solely on the Auto-Calibration Signal (ACS) region $y_{\text{acs}}$ via gradient descent:
\begin{equation}
\min_{\mathbf{G}} \left\|\mathbf{G} \ast y_{\text{acs}} - y_{\text{acs}} \right\|_2^2,
\end{equation}
subject to the zero-center constraint $\mathbf{G}[:, :, \kappa//2, \kappa//2] = 0$. In inference, the SPIRiT-calibrated k-space $y_{\text{spirit}}$ is estimated by applying the learned kernel $\mathbf{G}$ to $R$. The final pseudo-label $y_{\text{final}}$ is generated via a stochastic ensemble strategy:
\begin{equation}\label{eq:ensemble}
y_{\text{final}} = (1 - \mathcal{M}_{r}) \odot R_{t} + \mathcal{M}_{r} \odot y_{\text{spirit}}^{(t)},
\end{equation}
where $\mathcal{M}_{r}$ is a Bernoulli mask with sample rate $r$.

\subsection{Non-Local Self-Similarity Pixel Bank}
\label{sec:pixelbank}
To overcome supervision scarcity, we augment training by exploiting repetitive anatomical structures. For each location $(h,w)$ in the estimated image $\bar{x}=\mathcal{S}^{H} \mathcal{F}^{H}(y_{\text{final}})$, we extract a patch $p_{h,w} \in \mathbb{C}^{P \times P}$. Within a search window $\mathcal{N}_{h,w}$, we compute the cosine similarity with neighboring patches $p_i$:
\begin{equation}
    \cos\theta_{h,w,i} = \frac{\langle p_{h,w}, p_i \rangle}{\|p_{h,w}\|_2 \cdot \|p_i\|_2}.
\end{equation}
The pixel bank $B \in \mathbb{C}^{H \times W \times N}$ then stores the center pixels of the top-$N$ most similar patches:
\begin{equation}
    B[h,w,:] = \left[ p_{i_1}^{\text{cen}}, \dots, p_{i_N}^{\text{cen}} \right], \quad \text{with } \cos\theta_{h,w,i_1} \geq \dots \geq \cos\theta_{h,w,i_N}.
\end{equation}
This yields a non-local prior over feasible pixels, decoupled from the k-space sampling pattern $M_\Omega$.

\subsection{Loss Function}
\label{subsec:loss}
To learn from the NSS Pixel Bank, we transform the bank into the multi-coil k-space domain, denoted as $\bar{y}_k = \mathcal{F}\mathcal{S}(B[:,:,k])$ (where we set $N=K$ to ensure a one-to-one mapping). During each training step, this set of reference data is randomly shuffled along the similarity dimension, and each $\bar{y}_k$ is assigned to a training mask partition pair $(\Theta_k, \Lambda_k)$. To further prevent information leakage, we generate a regularization mask $\bar{M}_{\text{k}}$ strictly disjoint from the non-ACS part of the acquisition mask $M_{\Omega}$ (i.e., $\bar{M}_{\text{k}} \cap (M_{\Omega} \setminus \Omega_{\text{acs}}) = \emptyset$). We denote this strictly disjoint masking strategy as the Exclusive Mask. 

Lastly, we minimize a joint objective $\mathcal{L}_{\text{total}}$, which is formulated over the $K$ partitions as:
\begin{equation}
    \mathcal{L}_{\text{total}} = \mathcal{L}_{\text{dc}}+\lambda_{\text{reg}}\frac{1}{K} \sum_{k=1}^{K} \mathcal{L}\big(\bar{y}_k, (1-\bar{M}_{\text{k}})\odot f(\bar{y}_k \odot \bar{M}_{\text{k}}, \bar{M}_{\text{k}};\theta)\big),
\end{equation}
where $\lambda_{\text{reg}}$ balances data fidelity and non-local prior consistency.
\begin{figure}[t]
    \includegraphics[width=1\textwidth]{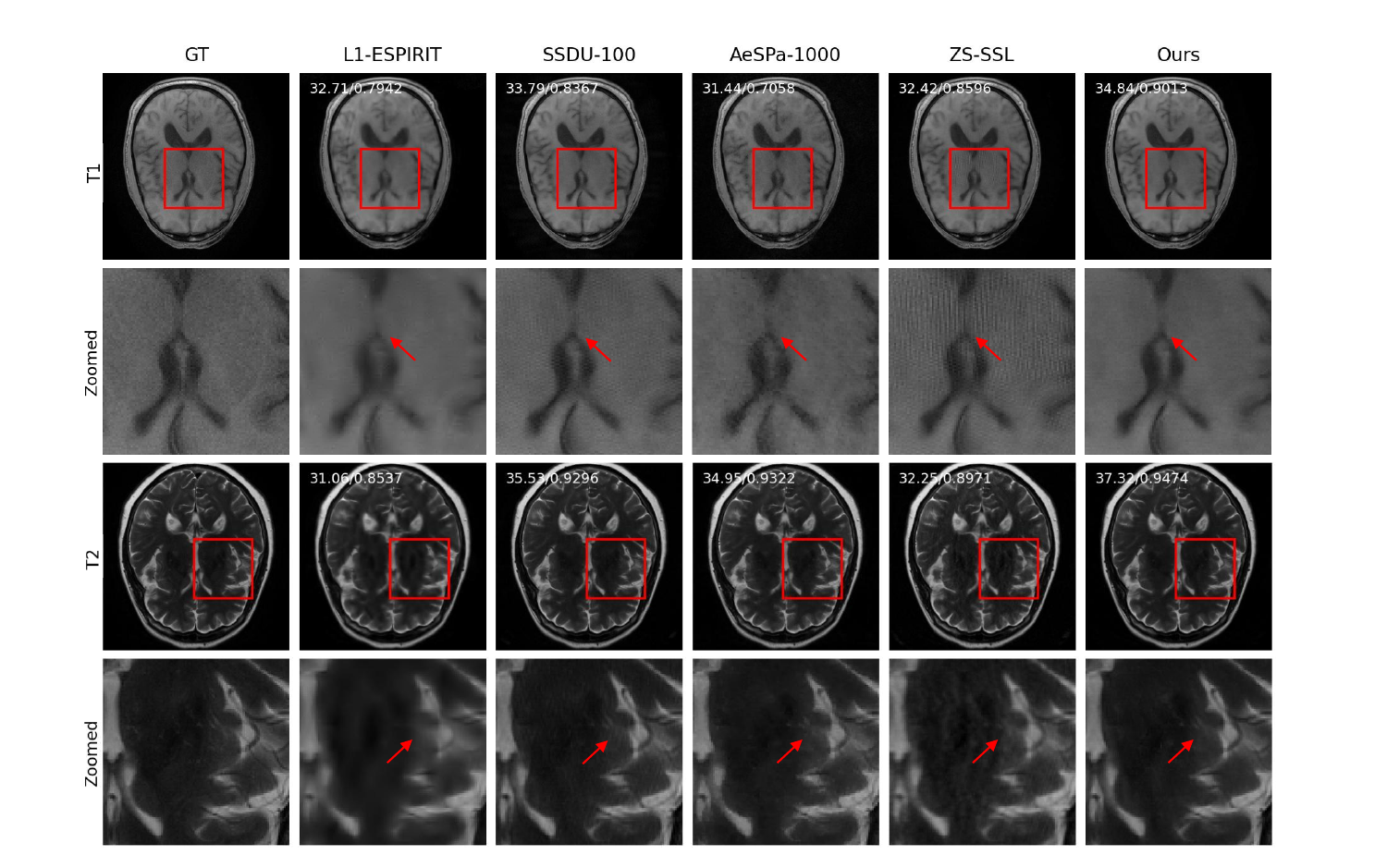}    
    \caption{Visualization of reconstruction results for the FastMRI brain (T1/T2) dataset with a 4$\times$ uniform mask. The red boxes indicate the corresponding zoomed-in regions, and the red arrows highlight the regions of interest.}
    \label{fig:T1}
\end{figure}
\section{Experiments}
\subsection{Dataset}
We evaluated the proposed method on the multi-coil brain (T1/T2) and knee subsets from the publicly available FastMRI dataset \cite{knoll2020fastmri}. We selected 20 slices for each subset. The coil sensitivity maps were estimated from the auto-calibration signal (ACS) using the ESPIRiT \cite{uecker2014espirit} algorithm. The reconstruction quality is evaluated using Peak Signal-to-Noise Ratio (PSNR) and Structural Similarity Index Measure (SSIM).
\subsection{Implementation details}
We compared our method with both numerical and DL methods: L1-ESPIRiT \cite{uecker2014espirit}, SSDU \cite{ssdu}, ZS-SSL \cite{zs-ssl}, and AeSPa \cite{joo2025aespa}. We adopted the unrolled denoising architecture \cite{liang2020deep} with 8 residual blocks (RBs). To ensure a fair comparison, the ZS-SSL baseline was evaluated using this identical network specification. We set $r=0.5$, $\lambda_{\text{reg}} = 0.1$, and $N=K=25$, while other hyperparameters were set the same as ZS-SSL. Since SSDU and AeSPa lack stopping criteria, we report both their average performance at peak PSNR during training (denoted by $^*$) and their final results at fixed iterations (100 epochs for SSDU and 1000 epochs for AeSPa, denoted as "-100" and "-1000") to reflect practical convergence.
\begin{table}[t]
\centering
\caption{Quantitative comparison (PSNR/SSIM) on the \textbf{FastMRI Brain T1} dataset. Best results are highlighted in bold.}
\label{tab:t1_results}
\footnotesize
\begin{tabular}{lcccccccc}
\toprule
\multirow{3}{*}{\textbf{Method}} & \multicolumn{4}{c}{\textbf{Uniform Mask}} & \multicolumn{4}{c}{\textbf{Variable Density Mask}} \\
\cmidrule(lr){2-5} \cmidrule(lr){6-9}
 & \multicolumn{2}{c}{\textbf{4x}} & \multicolumn{2}{c}{\textbf{8x}} & \multicolumn{2}{c}{\textbf{4x}} & \multicolumn{2}{c}{\textbf{8x}} \\
\cmidrule(lr){2-3} \cmidrule(lr){4-5} \cmidrule(lr){6-7} \cmidrule(lr){8-9} 
 & PSNR & SSIM & PSNR & SSIM & PSNR & SSIM & PSNR & SSIM \\
\midrule
L1-ESPIRiT      & 31.69 & 0.8163 & 23.40 & 0.6815 & 32.98 & 0.8264 & 26.57 & 0.7584 \\
SSDU-100        & 35.51 & 0.8775 & 29.34 & 0.8096 & 33.53 & 0.8253 & 32.97 & 0.8748 \\
AeSPa-1000      & 32.84 & 0.7919 & 26.56 & 0.7465 & 33.21 & 0.8241 & 29.04 & 0.7857 \\
ZS-SSL          & 36.03 & 0.9085 & 27.20 & 0.7261 & 35.87 & 0.8976 & 32.51 & 0.8774 \\
\textbf{Ours}   & \textbf{37.04} & \textbf{0.9265} & \textbf{29.46} & \textbf{0.8307} & \textbf{37.16} & \textbf{0.9284} & \textbf{33.66} & \textbf{0.8964} \\
\midrule
$\text{SSDU}^*$ & 36.27 & 0.9116 & 29.47 & 0.8177 & 36.32 & 0.9160 & 33.15 & 0.8852 \\ 
$\text{AeSPa}^*$& 34.74 & 0.8374 & 28.05 & 0.7391 & 35.23 & 0.8471 & 31.39 & 0.7872 \\
\bottomrule
\end{tabular}
\end{table}

\begin{table}[t]
\centering
\caption{Quantitative comparison (PSNR/SSIM) on the \textbf{FastMRI Brain T2} dataset. Best results are highlighted in bold.}
\label{tab:t2_results}
\footnotesize
\begin{tabular}{lcccccccc}
\toprule
\multirow{3}{*}{\textbf{Method}} & \multicolumn{4}{c}{\textbf{Uniform Mask}} & \multicolumn{4}{c}{\textbf{Variable Density Mask}} \\
\cmidrule(lr){2-5} \cmidrule(lr){6-9}
 & \multicolumn{2}{c}{\textbf{4x}} & \multicolumn{2}{c}{\textbf{8x}} & \multicolumn{2}{c}{\textbf{4x}} & \multicolumn{2}{c}{\textbf{8x}} \\
\cmidrule(lr){2-3} \cmidrule(lr){4-5} \cmidrule(lr){6-7} \cmidrule(lr){8-9} 
 & PSNR & SSIM & PSNR & SSIM & PSNR & SSIM & PSNR & SSIM \\
\midrule
L1-ESPIRiT & 30.62 & 0.8494 & 22.86 & 0.6514 & 32.44 & 0.8660 & 26.43 & 0.7683 \\
SSDU-100 &36.02 &0.9314 &28.78 &0.8075 &35.57 &0.9332 &31.87 &0.8870 
 \\
 AeSPa-1000 &32.51 &0.8601 &22.91 &0.6854 &32.18 &0.8594 &27.17 &0.8132 \\
ZS-SSL & 36.48 & 0.9403 & 26.88 & 0.7600 &35.21 &0.9285  &30.63 &0.8705 
 \\
\textbf{Ours} & \textbf{37.12} & \textbf{0.9452} & \textbf{28.79} & \textbf{0.8228} & \textbf{36.50} & \textbf{0.9467} & \textbf{32.38} & \textbf{0.9010} \\
\midrule
$\text{SSDU}^*$ &36.14 &0.9354 &28.98 &0.8183 &35.80 &0.9394&31.98 &0.8904\\
 $\text{AeSPa}^*$ &34.69 &0.8262 &31.06	&0.7533 &35.96 &0.8662 &31.79 &0.7882\\
\bottomrule
\end{tabular}%
\end{table}

\begin{table}[t]
\centering
\caption{Quantitative comparison (PSNR/SSIM) on the FastMRI Knee dataset. Best results are highlighted in bold.}
\label{tab:knee_results}
\footnotesize
\begin{tabular}{lcccccccc}
\toprule
\multirow{3}{*}{\textbf{Method}} & \multicolumn{4}{c}{\textbf{Uniform Mask}} & \multicolumn{4}{c}{\textbf{Variable Density Mask}} \\
\cmidrule(lr){2-5} \cmidrule(lr){6-9}
 & \multicolumn{2}{c}{\textbf{4x}} & \multicolumn{2}{c}{\textbf{8x}} & \multicolumn{2}{c}{\textbf{4x}} & \multicolumn{2}{c}{\textbf{8x}} \\
\cmidrule(lr){2-3} \cmidrule(lr){4-5} \cmidrule(lr){6-7} \cmidrule(lr){8-9} 
 & PSNR & SSIM & PSNR & SSIM & PSNR & SSIM & PSNR & SSIM \\
\midrule
L1-ESPIRiT &29.89 &0.8085 &23.66 &0.6912 &33.93 &0.8402 &24.84 &0.7118  \\
SSDU-100 &35.18 &0.8581 &31.29 &0.7818 &32.41 &0.7337 &31.50 &0.7693\\
AeSPa-1000 &31.21 &0.7251 &28.52 &0.7133 &32.24 &0.7464 &28.69 &0.7045\\
ZS-SSL &35.65 &0.8804 &30.12 &0.7799 &35.75 &0.8500 &31.05 &0.8118 \\
\textbf{Ours} &\textbf{35.90} &\textbf{0.8925} &\textbf{31.22} &\textbf{0.8293} &\textbf{37.16} &\textbf{0.9064} &\textbf{31.99} &\textbf{0.8409} \\
\midrule
 $\text{SSDU}^*$ &36.00 	&0.8756 &31.88 &0.7957 &36.61 &0.8868 &32.53 &0.8156 \\
 $\text{AeSPa}^*$ &34.69 &0.8954 &25.73 &0.7222 &33.64 &0.8910 &29.59 &0.8081\\
\bottomrule
\end{tabular}
\end{table}

\begin{table}[t]
    \centering
    \caption{Ablation studies on the FastMRI brain T1 dataset. Left: Component ablation. "Repo." denotes Dynamic Repository, "PB" denotes NSS Pixel Bank, and "EM" denotes Exclusive Mask. Right: Sensitivity analysis of the Bernoulli mask sampling ratio $r$.}
    \label{tab:ablations_combined}
    \footnotesize
    \begin{tabular}{lcccccc | ccc}
    \toprule
    \multicolumn{7}{c|}{\textbf{Components}} & \multicolumn{3}{c}{\textbf{Sampling Ratio ($r$)}} \\
    \cmidrule{1-7} \cmidrule{8-10}
    \textbf{Method} & \textbf{Repo.} & \textbf{SPIRiT} & \textbf{PB} & \textbf{EM} & \textbf{PSNR} & \textbf{SSIM} & \textbf{Ratio} & \textbf{PSNR} & \textbf{SSIM} \\
    \midrule
    Baseline     & $\times$   & $\times$   & $\times$   & $\times$   & 36.03 & 0.9085 & 0.3          & 36.88 & 0.9255 \\
    + Repo.      & \checkmark & $\times$   & $\times$   & \checkmark & 36.21 & 0.9101 & 0.4          & 36.89 & 0.9252 \\
    + SPIRiT     & $\times$   & \checkmark & $\times$   & \checkmark & 36.74 & 0.9174 & \textbf{0.5} & \textbf{37.04} & \textbf{0.9265} \\
    + Pixel Bank & $\times$   & $\times$   & \checkmark & \checkmark & 36.92 & 0.9252 & 0.6          & 36.94 & 0.9258 \\
    w/o EM       & \checkmark & \checkmark & \checkmark & $\times$   & 36.89 & 0.9258 & 0.7          & 36.92 & 0.9252 \\
    \textbf{Proposed} & \checkmark & \checkmark & \checkmark & \checkmark & \textbf{37.04} & \textbf{0.9265} &0.8  &36.84  &0.9230  \\
    \bottomrule
    \end{tabular}
\end{table}

\section{Results}
\subsection{Comparison with baseline methods}
Quantitative evaluations on the three datasets demonstrate the consistent superiority of our proposed method. As shown in Tables \ref{tab:t1_results}, \ref{tab:t2_results}, and \ref{tab:knee_results}, our method outperforms state-of-the-art SSL and ZS-SSL baselines (including SSDU, AeSPa, and ZS-SSL) across all acceleration factors and sampling masks. Notably, even under the extreme acceleration factor of $8\times$, our approach exhibits substantial PSNR gains over the baselines. 

Visual results are presented in Fig. \ref{fig:T1}. Baseline methods like L1-ESPIRiT tend to produce over-smoothed results, whereas SSDU and ZS-SSL suffer from significant artifacts. AeSPa struggles to recover high-frequency details, resulting in visible blurring across structural boundaries. In contrast, benefiting from the proposed physics-driven constraints and non-local priors, our method outperforms the baselines in artifact suppression and noise removal.
\subsection{Ablation study}
\subsubsection{Impact of components.}
As shown in Table \ref{tab:ablations_combined}, all proposed components consistently contribute to the final reconstruction quality. Starting from a baseline PSNR of 36.03 dB, introducing the CSM-Guided Dynamic Repository yields a stable improvement (+0.18 dB), indicating its effectiveness in filtering physically inconsistent artifacts during early training stages. Applying the SPIRiT-based regularization brings a more substantial gain (+0.71 dB), which validates the importance of explicitly enforcing multi-coil k-space consistency. Among individual modules, the NSS Pixel Bank achieves the most significant performance boost, improving PSNR by 0.89 dB (reaching 36.92 dB) over the baseline. This highlights that exploiting image-domain self-similarity is highly effective in compensating for the severe supervision scarcity inherent in single-scan ZS-SSL. Crucially, simply integrating all modules without the Exclusive Mask (w/o EM) results in a performance drop to 36.89 dB, which is even lower than using the Pixel Bank alone. This degradation occurs because directly reconstructing pseudo-labels without strict masking leads to severe information leakage.

\subsubsection{Impact of Bernoulli Sampling Ratio ($r$).}
Table \ref{tab:ablations_combined} shows that performance peaks at $r=0.5$ (37.04 dB). 
Specifically, lower ratios (e.g., $r \le 0.4$) tend to over-rely on the historical predictions within the dynamic repository, which impedes the recovery of fine details. Conversely, higher ratios (e.g., $r \ge 0.6$) excessively prioritize the SPIRiT estimates, thereby increasing the network's vulnerability to ACS calibration errors. This indicates that a balanced fusion of repository stability and SPIRiT consistency optimally mitigates kernel artifacts. 
Consequently, we adopt $r=0.5$ for all experiments.
\section{Conclusion}
In this work, we presented a physics-driven framework to address the challenges of supervision scarcity and optimization instability inherent in ZS-SSL for MRI reconstruction. By synergizing a CSM-Guided Dynamic Repository with SPIRiT-based regularization, our method enforces k-space physical consistency while dynamically filtering artifacts. Moreover, the introduction of an NSS Pixel Bank effectively augments supervision by explicitly mining repetitive anatomical structures within the image domain. Extensive experiments on the FastMRI dataset demonstrate that our approach achieves state-of-the-art performance, bridging the performance gap between zero-shot learning and fully supervised methods.
%
%
%
\bibliographystyle{splncs04}
\bibliography{mybib}

@inproceedings{zs-ssl,
title={Zero-Shot Self-Supervised Learning for {MRI} Reconstruction},
author={Burhaneddin Yaman and Seyed Amir Hossein Hosseini and Mehmet Akcakaya},
booktitle={International Conference on Learning Representations},
year={2022},
url={https://openreview.net/forum?id=085y6YPaYjP}
}

@article{spicer,
  title={SPICER: Self-supervised learning for MRI with automatic coil sensitivity estimation and reconstruction},
  author={Hu, Yuyang and Gan, Weijie and Ying, Chunwei and Wang, Tongyao and Eldeniz, Cihat and Liu, Jiaming and Chen, Yasheng and An, Hongyu and Kamilov, Ulugbek S},
  journal={Magnetic resonance in medicine},
  volume={92},
  number={3},
  pages={1048--1063},
  year={2024},
  publisher={Wiley Online Library}
}

@article{knoll2020fastmri,
  title={fastMRI: A publicly available raw k-space and DICOM dataset of knee images for accelerated MR image reconstruction using machine learning},
  author={Knoll, Florian and Zbontar, Jure and Sriram, Anuroop and Muckley, Matthew J and Bruno, Mary and Defazio, Aaron and Parente, Marc and Geras, Krzysztof J and Katsnelson, Joe and Chandarana, Hersh and others},
  journal={Radiology: Artificial Intelligence},
  volume={2},
  number={1},
  pages={e190007},
  year={2020},
  publisher={Radiological Society of North America}
}

@article{uecker2014espirit,
  title={ESPIRiT—an eigenvalue approach to autocalibrating parallel MRI: where SENSE meets GRAPPA},
  author={Uecker, Martin and Lai, Peng and Murphy, Mark J and Virtue, Patrick and Elad, Michael and Pauly, John M and Vasanawala, Shreyas S and Lustig, Michael},
  journal={Magnetic resonance in medicine},
  volume={71},
  number={3},
  pages={990--1001},
  year={2014},
  publisher={Wiley Online Library}
}

@article{lustig2010spirit,
  title={SPIRiT: iterative self-consistent parallel imaging reconstruction from arbitrary k-space},
  author={Lustig, Michael and Pauly, John M},
  journal={Magnetic resonance in medicine},
  volume={64},
  number={2},
  pages={457--471},
  year={2010},
  publisher={Wiley Online Library}
}

@article{pruessmann1999sense,
  title={SENSE: sensitivity encoding for fast MRI},
  author={Pruessmann, Klaas P and Weiger, Markus and Scheidegger, Markus B and Boesiger, Peter},
  journal={Magnetic Resonance in Medicine: An Official Journal of the International Society for Magnetic Resonance in Medicine},
  volume={42},
  number={5},
  pages={952--962},
  year={1999},
  publisher={Wiley Online Library}
}

@article{ssdu,
  title={Self-supervised learning of physics-guided reconstruction neural networks without fully sampled reference data},
  author={Yaman, Burhaneddin and Hosseini, Seyed Amir Hossein and Moeller, Steen and Ellermann, Jutta and U{\u{g}}urbil, K{\^a}mil and Ak{\c{c}}akaya, Mehmet},
  journal={Magnetic resonance in medicine},
  volume={84},
  number={6},
  pages={3172--3191},
  year={2020},
  publisher={Wiley Online Library}
}

@inproceedings{joo2025aespa,
  title={AeSPa: Attention-guided Self-supervised Parallel Imaging for MRI Reconstruction},
  author={Joo, Jinho and Kim, Hyeseong and Won, Hyeyeon and Lee, Deukhee and Eo, Taejoon and Hwang, Dosik},
  booktitle={Proceedings of the Computer Vision and Pattern Recognition Conference},
  pages={5217--5226},
  year={2025}
}

@article{ma2025pixel2pixel,
  author       = {Qing Ma and
                  Junjun Jiang and
                  Xiong Zhou and
                  Pengwei Liang and
                  Xianming Liu and
                  Jiayi Ma},
  title        = {Pixel2Pixel: {A} Pixelwise Approach for Zero-Shot Single Image Denoising},
  journal      = {IEEE Transactions on Pattern Analysis and Machine Intelligence},
  volume       = {47},
  number       = {6},
  pages        = {4614--4629},
  year         = {2025}
}

@inproceedings{huang2021neighbor2neighbor,
  title={Neighbor2neighbor: Self-supervised denoising from single noisy images},
  author={Huang, Tao and Li, Songjiang and Jia, Xu and Lu, Huchuan and Liu, Jianzhuang},
  booktitle={Proceedings of the IEEE/CVF conference on computer vision and pattern recognition},
  pages={14781--14790},
  year={2021}
}

@inproceedings{mansour2023zero,
  title={Zero-shot noise2noise: Efficient image denoising without any data},
  author={Mansour, Youssef and Heckel, Reinhard},
  booktitle={Proceedings of the IEEE/CVF Conference on Computer Vision and Pattern Recognition},
  pages={14018--14027},
  year={2023}
}

@article{liang2020deep,
  title={Deep magnetic resonance image reconstruction: Inverse problems meet neural networks},
  author={Liang, Dong and Cheng, Jing and Ke, Ziwen and Ying, Leslie},
  journal={IEEE Signal Processing Magazine},
  volume={37},
  number={1},
  pages={141--151},
  year={2020},
  publisher={IEEE}
}
%
\end{document}